# Lossless Attention in Convolutional Networks for Facial Expression Recognition in the Wild

Chuang Wang[1], Ruimin Hu[1], Min Hu[1], Jiang Liu[1],
Ting Ren[2], Shan He[2], Ming Jiang[2], Jing Miao[2]

[1] National Engineering Research Center for Multimedia Software (NERCMS), Wuhan University, Wuhan, China,
[2] Iflytek Xingzhi Technology Co., Ltd, Wuhan, China

*Abstract*—Unlike the constraint frontal face condition, faces in the wild have various unconstrained interference factors, such as complex illumination, changing perspective and various occlusions. Facial expressions recognition (FER) in the wild is a challenging task and existing methods can't perform well. However, for occluded faces (containing occlusion caused by other objects and self-occlusion caused by head posture changes), the attention mechanism has the ability to focus on the non-occluded regions automatically. In this paper, we propose a Lossless Attention Model (LLAM) for convolutional neural networks (CNN) to extract attention-aware features from faces. Our module avoids decay information in the process of generating attention maps by using the information of the previous layer and not reducing the dimensionality. Sequentially, we adaptively refine the feature responses by fusing the attention map with the feature map. We participate in the seven basic expression classification sub-challenges of FG-2020 Affective Behavior Analysis in-the-wild Challenge. And we validate our method on the Aff-Wild2 datasets released by the Challenge. The total accuracy (Accuracy) and the unweighted mean (F1) of our method on the validation set are 0.49 and 0.38 respectively, and the final result is 0.42 (0.67 F1_Score + 0.33 Accuracy).

*Keywords-Facial expression recognition; Attention mechanism; Avoiding information loss; Convolutional neural networks*

## I. INTRODUCTION

Facial expression has practical value in the fields of human-computer interaction, such like human behavior understanding, mental health assessment, etc. Many methods of automatically recognizing facial expression have been proposed in the literature. Compared with the task of FER under laboratory-controlled conditions [17] [18] [19], FER in-the-wild condition is more difficult because of the more interference factors, such as complex illumination, changing perspective and posture, and various occlusions.

As traditional methods, hand-crafted features have been wildly used for FER [1], such as Gabor wavelet coefficients [12], local binary pattern (LBP) [13] and histograms of oriented gradients (HOG) [14]. However, with the growth of training data in FER and the excellent representing ability of deep neural networks (DNNs), many deep learning methods have been proposed. The special hierarchical structure of convolution neural networks can extract informative and discriminative feature. Motivated by this. Mollahosseini proposed a deep convolution neural network which stacked two convolutional layers with four Inception layers and the model gained comparable results [2]. Li proposed Deep Locality-Preserving CNN (DLP-CNN) to recognize facial expressions in the wild [3], The disadvantage of the above-mentioned methods is that the features they extracted are less-discriminative because of equal weights for informative regions (or channels) and non-informative regions (or channels). Motivated by the perspective that not all regions and channels contain useful information for expression recognition [39] [41], we think that attention mechanism is an effective solution to highlight useful region and suppress other regions' interferences.

Human can rapidly orientate towards salient objects in a cluttered visual scene [4] [5], which is the attention mechanism of our visual perception [6]. Recently, attention mechanism has been successfully applied in many computer vision tasks [5], such as visual question answering [7], image caption [8] and fine-grained image recognition [9]. Zhao estimated multiple 2-dimensional attention maps for person re-Identification to emphasize more informative regions and suppress less useful ones [11]. Attention mechanisms are also applied to facial expression recognition tasks. For instance, to solve occlusion and pose problems in FER, Wang proposed a *Region Attention Network* (RAN), which adaptively captured the important facial regions [40]. However, the computational complexity of the RAN increases because of the extra facial region decomposition procedure.

Different with [40], we innovatively propose an attention module for CNNs which directly takes the entire face image as input without decomposition procedure. What's more, it can adaptively focus attention on the more discerning face area and generate attention features. In the previous attention methods, in order to reduce the calculation amount or increase the receptive field to capturing long range interactions, dimensionality reduction was performed in the process of generating attention maps. However, it will cause information loss and has a negative effect on recognition accuracy [34]. Simultaneously, the previous attention-based methods generate attention map only from the current layer while ignore some the useful information contained by previous layers. The useful information reduction layer by layer will lead to information loss. In order to avoid this disadvantage, we generate attention map using previous and current features information and furtherly keep the dimension of attention map as same as the size of original features for preserving more information. Our approach can generate more precise attention maps and adaptively focus on strong discernment area of facial expressions.

## II. RELATED WORK

In this section, we first survey deep learning methods for FER and then discuss the application of attention mechanism in CNN

### A. Deep Learning Methods for FER

Nowadays, the FER is transiting from laboratory-controlled to in-the-wild conditions. The facial images, captured in laboratory-controlled condition, are frontal, non-occlusive and with same illumination. Thus, it is less-challenging for FER. [20] [21] [22] [23] have performed well by directly using deep features for FER in laboratory-controlled condition.

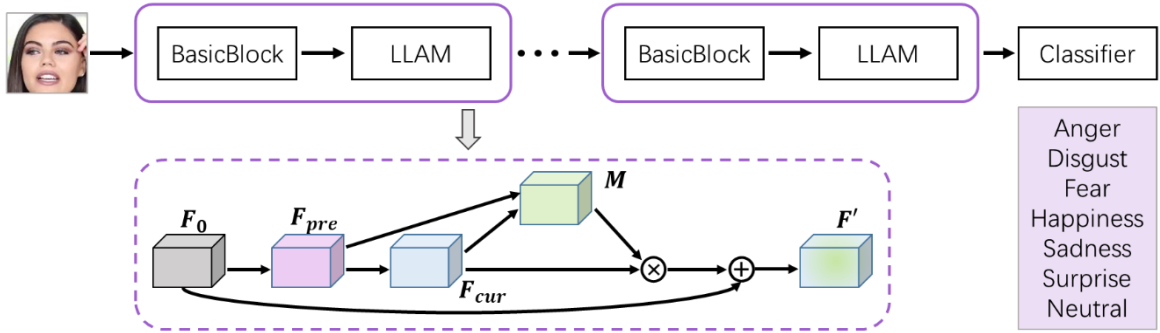

Fig. 1. Overall structure of the proposed LLA-Net. The Network is built by stacking modules which consist of BasicBlock from ResNet and our LLAM. The structure of the combined module is shown in the dotted box below, where $F_0$ denotes the output of the last combined module. The details of LLAM will be introduced in next section.

However, when facial images are captured in the wild, the illumination, posture and occlusions can vary significantly, which makes it harder to achieve high-accuracy for FER. Acharya proposed using a convolutional network where the traditional pooling layer is replaced by covariance pooling for expression recognition. The second-order statistics covariance pooling able to capture distortions in regional facial features better than first-order statistics [28]. Li collected a large-scale facial expression dataset RAF-DB in the wild for unconstrained FER [3]. In addition, they also proposed a DLP-CNN by add LP loss to the fundamental convolutional architecture to address the ambiguity and multi-modality of real-world expressions. Georgescu combined features learned by CNN and handcrafted features obtained from BOVW and achieved superior results on datasets in the wild [30]. Liu proposed a MPCNN for multi-view FER [31]. This approach is able to avoid overfitting and robust to large head pose variances. However, not all face areas contribute equally to expressions, and local area contribution will be greater on in-the-wild faces. Because CNN lacks this ability to distinguish important regions, attention can make up for this deficiency.

### B. Attention Mechanism in CNN

The attention mechanism has been widely applied in computer vision tasks. Hu developed a Squeeze-and-Excitation (SE) block to model the relationship of channels and used the global average-pooled features to calculate the attention level per channel [33]. The effectiveness of SE model in large-scale image classification and object detection tasks inspired Wu, who just considered the relationship between adjacent channels for achieving high-efficiency [34]. Xu proposed a Spatial Memory Network for Visual Question Answering (SMem-VQA), which is explicit spatial attention[35]. Meng designed a Frame Attention Networks (FAN) for video-based FER. FAN learned self-attention weights and relation-attention weights to assess the importance of frames [38]. Li cropped facial regions and assigned attention weights for 24 patches adaptively [5]. Channel attention mechanism adaptively assigns importance to channels, but local regions within each channel are not explicitly emphasized or suppressed. Similarly, spatial attention treats each channel completely equally, ignoring the differences of features between channels. However, combining two types of attention, i.e., mixed attention, can produce complementary effects. Zhu designed an attention module to generate for the feature maps. The soft 3D masks efficiently highlighted facial expression sensitive areas [39]. Similar to [39], we use mixed attention to focus on channel and spatial differences. The above-mentioned methods only generate the attention maps using the information from current layer and some of them even perform dimensionality reduction. Different from them, our approach reuses the output of the previous layer and abandon dimensionality reduction, which effectively avoids information loss and generates more reliable attention maps.

## III. APPROACH

### A. Overview

Figure 1 shows the outline of our proposed framework *Lossless Attention Convolutional Network* (LLA-Net). We embed our LLAM into the intermediate layer of the convolutional neural network to generate attention maps to refine feature maps generated by the convolutional layers. Due to the efficient and excellent performance of ResNet-18 in computer vision tasks, we select ResNet-18 as our Backbone network. We combine the proposed LLAM with BasicBlock in the network. At the end of the network, a fully connected layer provides a mapping operation from feature space to classification space before the final softmax layer predicts the emotion classes. Different from the *Convolutional Block Attention Module*（CBAM）[37], which generates spatial and channel attention separately, our model directly computes a 3D attention map which combines spatial and channel attentions.

### B. Lossless Attention Module

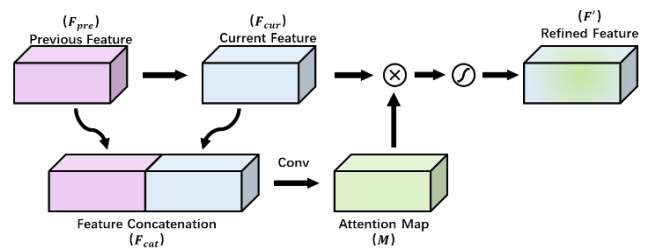

Fig. 2. Structure of the proposed LLAM. The feature map $F_{pre}$ from previous convolution layer and feature map $F_{cur}$ from current convolution layer are connected to form $F_{cat}$, and then the attention map $M$ is obtained after a convolution layer. The element-wise multiplication is performed between $F_{cur}$ and $M$, and then pass an Sigmoid function to get the final refined feature map $F'$.

The finite parameter quantity of convolutional layer limits its feature extraction capability. Thus, some useful information in previous layer is abandoned. In order to fully employ the information in previous layer, we reuse the previous feature map to generate an attention map. What's more, inspired by [34], which proposed that dimensionality reduction has a negative effect on recognition accuracy when generating attention maps. We don't conduct dimensionality reduction same as [33] [36] [37].

TABLE I. Image distribution of seven basic classes in Aff-Wild2.

| Subset | Anger | Disgust | Fear | Happy | Sad | Surprise | Neutral | **Total** |
|---|---|---|---|---|---|---|---|---|
| Training | 25634 | 11490 | 19279 | 171902 | 102934 | 43306 | 546039 | **920584** |
| Validation | 7389 | 7699 | 18216 | 61307 | 35334 | 23710 | 174827 | **328482** |

TABLE II. Distribution of under sampled and supplemented training set of Aff-Wild2. For categories with two numbers, the value below is the number of extra samples from other dataset (AffectNet or RAF-DB).

| Subset | Anger | Disgust | Fear | Happy | Sad | Surprise | Neutral | **Total** |
|---|---|---|---|---|---|---|---|---|
| Training | 25634 / 24242 | 11490 / 5062 | 19279 / 6192 | 86164 | 102934 | 43306 | 46073 | **370376** |

Figure 2 shows the structure of LLAM. $F_{pre} \in \mathbb{R}^{C \times H \times W}$ and $F_{cur} \in \mathbb{R}^{C \times H \times W}$ are the feature maps, generated by the previous and current convolution layers, respectively. $C$, $H$ and $W$ represent the number of channels, height and width of the feature map, respectively. In order to reuse the information from the previous layer, we concatenate previous and current feature maps, denoted as $F_{cat} \in \mathbb{R}^{2C \times H \times W}$. Then a standard convolution operation is used to generate the attention map to capture cross-channel and cross-spatial interaction. The whole attention process can be summarized as:

$$F_{cat} = [F_{pre} : F_{cur}],$$
$$M = \sigma(f(F_{cat})) = \sigma(f([F_{pre} : F_{cur}])),$$
$$F' = F_{cur} \otimes M,$$

where $f$ represents the convolution layer, $\sigma$ denotes the Sigmoid function, $M \in \mathbb{R}^{C \times H \times W}$ is the generated attention map, $F'$ denotes the refined feature map, and $\otimes$ represents element-wise multiplication. It should be noted that we reduce the channel number from $2C$ to $C$ when we generate attention maps $M$ from $F_{cat}$ for keeping the size consistent with the original feature maps but we do not reduce the dimension of attention maps.

## IV. EXPERIMIENTS

### A. Dataset and Preprocessing

***Aff-Wild2.*** Aff-Wild2 [42] [43] is an extension of the Aff-Wild [44] [45] [36] [32] database. As a large scale database, Aff-Wild2 contains 564 videos of around 2.8M frames (the largest existing one). Aff-Wild2 is the first comprehensive in-the-wild database to contain annotations for all 3 behavior tasks, i.e., valence-arousal estimation, seven basic expression classification and facial action unit detection. In terms of annotations, 561 videos have annotations for valence-arousal, 63 videos have annotations for 8 AUs and 547 videos have annotations for the 7 basic expressions. We participate in the seven basic expression classification sub-challenge, so we use the subset with seven expressions labels (Anger, Disgust, Fear, Happiness, Sadness, Surprise and Neutral). For this sub-challenge, the cropped and aligned images provided by the organizer contain two subsets for training and validation. The detail information is shown in Table I.

***Preprocessing.*** We use images that the organizer has cropped and aligned into a size of 112 x 112. As can be seen from Table I, the original dataset is seriously class-imbalanced. The original dataset includes each frame in the expression change process, and some examples are shown in the top row of Figure 3. Obviously, because the expressions change gradually, the frames in adjacent are extremely similar and information is redundant. Class-imbalance will seriously affect the performance of the classifier and cause the classification results to be biased to the categories with a large number of samples. Therefore, a method is required to make the number of seven categories samples equal.

Based on the above analysis, we adopt the following methods to alleviate the problem of class-imbalance: i) We under sample two categories (neutral and happy) with the largest sample size. Specifically, in the continuous frame sequence of the same person and expression, we sample one every $k$ frames, and the $k$ values of neutral and happy are set to 12 and 2 respectively. ii) Although we under sample neutral and happy, the samples of anger, disgust and fear are still much less than that of other categories. Therefore, we choose some samples of corresponding categories from the AffectNet dataset and add them into these three categories. Specially, as the sample size of Disgust is still small, we have added extra corresponding samples of the RAF dataset to this class. The distribution of the final training set is shown in Table II.

The original images are cropped and aligned by the organizer contain too much background, which will introduce too much noise and adversely affect the result of expression recognition. To solve this problem, we crop the image again by [29]. Some examples of the cropped images are shown in the bottom row of Figure 3. During training, data augmentation techniques, i.e., random horizontal flip and random cropping are adopted. In testing phase, TenCrop technology is applied.

TABLE III. Test results of the proposed method on the validation set of the Aff-Wild2 dataset

|  | Anger | Disgust | Fear | Happy | Sad | Surprise | Neutral | Accuracy | F1 | Result | Baseline |
|---|---|---|---|---|---|---|---|---|---|---|---|
| result | 0.31 | 0.36 | 0.42 | 0.52 | 0.30 | 0.26 | 0.58 | 0.49 | 0.38 | **0.42** | 0.36 |

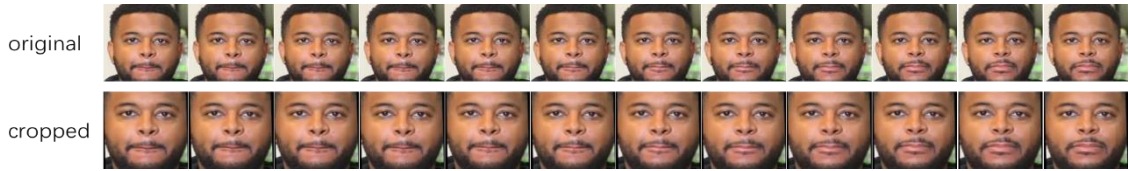

Fig. 3 Examples of original (top) and cropped (bottom) images in Aff-Wild2

*B. Implementation Details*

We use ResNet-18 as the main network. In order to fit the input size of 112 x 112, the first convolutional layer of the network does not change the size of the feature map. It should be noted that the ResNet-18 network we used was pre-trained on the RAF-DB [29] dataset.

We adopt batch-based SGD algorithm with a momentum of 0.9 and weight decay 5e-4 for model optimization. We set the initial learning rate to 0.01, and after 60 epochs, it is reduced by exponential decay method with a decay rate of 0.9. The batch size is set to 256 and training stopped after 60 epochs, then the model that performed best on the validation set was saved. We implement our method using Pytorch framework and all experiments are carried out on the Tesla V100-NVLINK with 32GB memory.

*C. Results*

For the convenience of description, we call the dataset cropped and aligned by the organizer for seven basic expression classification sub-challenges as OrigSet. We perform some processing on OrigSet. For the training set, class-balance and cropping as described are used. For the validation set, we only do the cropping operation. The experimental results and confusion matrix on the validation set are shown in Table I and Figure 3, respectively. The total accuracy (Accuracy) and the unweighted mean (F1) of our method on the validation set are 0.49 and 0.38 respectively, and the final result is 0.42 (0.67 F1_Score + 0.33 Accuracy), which is 0.07 higher than the baseline 0.36. The accuracy of each expression class is shown in Table III.

As can be seen from the confusion matrix in Figure 4, a large number of samples in each class are misclassified as Neutral but not the opposite. We use the following analyses to explain this phenomenon. Firstly, facial expressions change gradually, whose motion amplitudes are from weak to strong and then weak. Thus, many weak expressions are contained in the sequence of the expression images sequence. The weak expressions are similar with the Neutral expressions and easily confusing. Secondly, we collectively refer to the other classes except Neutral as *Others*, and refer to the strong expression features the weak expression features as $F_s$ and $F_w$ respectively. The model learns both $F_s$ and $F_w$ from *Others*, only $F_w$ can be learned from Neutral. If there is a weak or Neutral expression, its features are only $F_w$ and no $F_s$, and thus it is likely that this expression is recognized as a neutral expression. The above two analyses can explain the reason of the weak expressions of Others tending to be misclassified as neutral instead of the opposite.

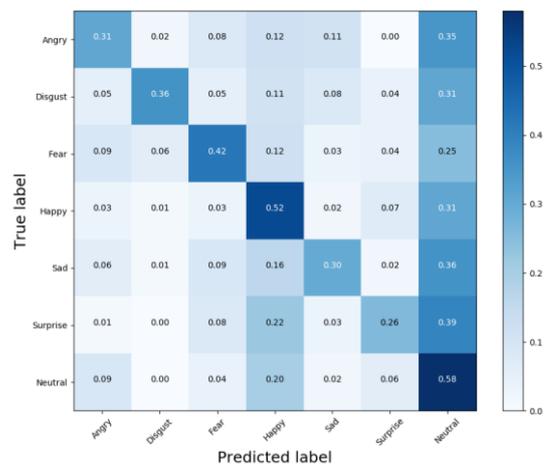

Fig. 4. The confusion matrix of seven basic expressions

## V. CONCLUSIONS

In this paper, we have proposed the Lossless attention module (LLAM), which is used in CNN for in-the-wild expression recognition. Our attention module makes our attention maps retain more useful information by reusing the feature maps from the previous layers and avoiding dimensionality reduction. We also class-balance the Aff-wild2 dataset and re-crop the original image to preserve less background. We participate in the *seven basic expression classification* sub-challenges of *FG-2020 Affective Behavior Analysis in-the-wild Challenge*. And we conduct experiments on Aff-Wild2 dataset and got result that higher than the baseline

In the future, we will apply our attention module to other convolutional neural networks and conduct experiments on other datasets. And we will continue to optimize our proposed LLAM.

REFERENCES

[1] Y. Chen, H. Hu, "Facial Expression Recognition by Inter-Class Relational Learning," IEEE Access, vol. 7, pp. 94106-94117, 2019.
[2] A. Mollahosseini, D. Chan, M. H. Mahoor, "Going deeper in facial expression recognition using deep neural networks,", *Proc. IEEE Winter Conf. Appl. Comput. Vis. (WACV)*, pp. 1-10, Mar. 2016.
[3] S. Li, W. Deng, J. P. Du, "Reliable crowdsourcing and deep locality-preserving learning for expression recognition in the

wild,", *Proceedings of the 2017 IEEE Conference on Computer Vision and Pattern Recognition*, pp. 2584-2593, 2017.

[4] L. Itti and C. Koch, "Computational modelling of visual attention,", *Nature Rev. Neurosci.*, vol. 2, no. 3, pp. 194–203, 2001.

[5] Y. Li, J. Zeng, S. Shan, X. Chen, "Occlusion aware facial expression recognition using CNN with attention mechanism,", *IEEE Trans. Image Process.*, vol. 28, no. 5, pp. 2439-2450, May 2019.

[6] Fei Wang, Mengqing Jiang, Chen Qian, Shuo Yang, Cheng Li, Honggang Zhang, Xiaogang Wang, and Xiaoou Tang, "Residual attention network for image classification,", arXiv preprint arXiv:1704.06904, 2017.

[7] C. Zhu, Y. Zhao, S. Huang, K. Tu, and Y. Ma, "Structured attentions for visual question answering,", in Proc. ICCV, 2017, pp. 1291–1300.

[8] K. Xu, J. Ba, R. Kiros, K. Cho, A. Courville, R. Salakhutdinov, R Zemel, Y. Bengio, "Show, attend and tell: Neural image caption generation with visual attention," in *Proc. ICML*, 2015, pp. 2048–2057.

[9] H. Zheng, J. Fu, T. Mei, and J. Luo, "Learning multi-attention convolutional neural network for fine-grained image recognition," in *Proc. ICCV*, Oct. 2017, pp. 5219–5227.

[10] L. Mou, X. X. Zhu, "Learning to Pay Attention on Spectral Domain: A Spectral Attention Module-Based Convolutional Network for Hyperspectral Image Classification,", *IEEE Trans. Geosci. Remote Sens.*, vol. 58, pp. 110 - 122, Jan. 2020.

[11] L. Zhao, X. Li, Y. Zhuang, and J. Wang, "Deeply-learned part-aligned representations for person re-identification," in *Proc. ICCV*, Oct. 2017, pp. 3239–3248.

[12] Y. Tian, T. Kanade, and J. F. Cohn, "Evaluation of Gabor-wavelet-based facial action unit recognition in image sequences of increasing complexity," in Proc. 5th IEEE Int. Conf. Autom. Face Gesture Recognit., May 2002, pp. 229_234.

[13] L. Zhong, Q. Liu, P. Yang, B. Liu, J. Huang, and D. N. Metaxas, "Learning active facial patches for expression analysis," in *Proc. IEEE Conf. Comput. Vis. Pattern Recognit.*, Jun. 2012, pp. 2562_2569.

[14] R. Girshick, J. Donahue, T. Darrell, and J. Malik, "Rich feature hierarchies for accurate object detection and semantic segmentation," in *Proc. IEEE Conf. Comput. Vis. Pattern Recognit.*, Jun. 2014, pp. 580_587.

[15] F. Schroff, D. Kalenichenko, and J. Philbin, "FaceNet: A unified embedding for face recognition and clustering," in Proc. IEEE Comput. Soc. Conf. Comput. Vis. Pattern Recognit., Jun. 2015, pp. 815–823.

[16] S. Ren, K. He, R. Girshick, J. Sun, "Faster R-CNN: Towards real-time object detection with region proposal networks," IEEE Trans. Pattern Anal. Mach. Intell., vol. 39, no. 6, pp. 1137-1149, Jun. 2017.

[17] P. Lucey, J. F. Cohn, T. Kanade, J. Saragih, Z. Ambadar, I. Matthews, "The Extended Cohn–Kanade dataset (CK+): A complete dataset for action unit and emotion-specified expression," *Proc. IEEE Comput. Soc. Conf. Comput. Vis. Pattern Recognit. Workshops*, pp. 94-101, Jun. 2010.

[18] G. Zhao, X. Huang, M. Taini, S. Z. Li, M. Pietikäinen, "Facial expression recognition from near-infrared videos," *Image and Vision Computing*, vol. 29, pp. 607-619, Aug. 2011.

[19] M. Pantic, M. Valstar, R. Rademaker, and L. Maat, "Web-based database for facial expression analysis," In Proc. IEEE Int'l Conf. Multmedia and Expo (ICME'05), 2005.

[20] H. Jung, S. Lee, J. Yim, S. Park, J. Kim, "Joint Fine-Tuning in Deep Neural Networks for Facial Expression Recognition," *2015 IEEE International Conference on Computer Vision (ICCV)* IEEE, 2015.

[21] X. Zhao, X. Liang, L. Liu, T. Li, Y. Han, N. Vasconcelos, S. Yan, "Peak-Piloted Deep Network for Facial Expression Recognition," *European Conference on Computer Vision* Springer, Cham, 2016.

[22] Majumder, Anima, L. Behera, and V. K. Subramanian, "Automatic Facial Expression Recognition System Using Deep Network-Based Data Fusion," *IEEE Transactions on Cybernetics*, PP.99(2016):1-12.

[23] S. Xie, H. Hu, "Facial expression recognition using hierarchical features with deep comprehensive multipatches aggregation convolutional neural networks," *IEEE Trans. Multimedia*, vol. 21, no. 1, pp. 211-220, Jan. 2019.

[24] Goodfellow I.J. et al. (2013) Challenges in Representation Learning: A Report on Three Machine Learning Contests. In: Lee M., Hirose A., Hou ZG., Kil R.M. (eds) Neural Information Processing. ICONIP 2013. Lecture Notes in Computer Science, vol 8228. Springer, Berlin, Heidelberg.

[25] Michael J. Lyons, Shigeru Akemastu, Miyuki Kamachi, Jiro Gyoba, "Coding Facial Expressions with Gabor Wavelets," *3rd IEEE International Conference on Automatic Face and Gesture Recognition*, pp. 200-205, 1998.

[26] Mollahosseini, Ali, B. Hasani, and M. H. Mahoor, "AffectNet: A Database for Facial Expression, Valence, and Arousal Computing in the Wild," *IEEE Transactions on Affective Computing* (2017):1-1.

[27] A. Dhall, R. Goecke, S. Lucey, and T. Gedeon. Static Facial Expression Analysis In Tough Conditions: Data, Evaluation Protocol And Benchmark. In ICCVW, BEFIT'11, 2011.

[28] D. Acharya, Z. Huang, D. Pani Paudel, L. Van Gool, "Covariance pooling for facial expression recognition", *IEEE Conference on Computer Vision and Pattern Recognition Workshops*, pp. 367-374, 2018.

[29] K. Zhang, Z. Zhang, Z. Li, Y. Qiao, "Joint Face Detection and Alignment Using Multitask Cascaded Convolutional Networks", IEEE Signal Processing Letters, vol. 23, no. 10, pp. 1499-1503, Oct. 2016.

[30] M. Georgescu, R. T. Ionescu, M. Popescu, "Local learning with deep and handcrafted features for facial expression recognition", *IEEE Access*, vol. 7, pp. 64 827-64 836, 2019.

[31] Y. Liu, J. Zeng, S. Shan, Z. Zheng, "Multi-Channel Pose-Aware Convolution Neural Networks for Multi-View Facial Expression Recognition", *IEEE FG 2018*, pp. 458-465, May 2018.

[32] D. Kollias, M. Nicolaou, I. Kotsia, G. Zhao, S. Zafeiriou, "Recognition of affect in the wild using deep neural networks", Proceedings of the IEEE Conference on Computer Vision and Pattern Recognition Workshop, 2017.

[33] J. Hu, L. Shen, S. Albanie, G. Sun, and E. Wu, "Squeeze-and-excitation networks," in Proc. IEEE Conf. Comput. Vis. Pattern Recognit. (CVPR), Jun. 2019, pp. 7132–7141.

[34] Q. Wang, B. Wu, P. Zhu, P. Li, W. Zuo, Q. Hu, "ECA-Net: Efficient Channel Attention for Deep Convolutional Neural Networks.", arXiv preprint arXiv:1910.03151, 2019.

[35] Xu H., Saenko K. (2016) Ask, Attend and Answer: Exploring Question-Guided Spatial Attention for Visual Question Answering. In: Leibe B., Matas J., Sebe N., Welling M. (eds) Computer Vision – ECCV 2016. ECCV 2016. Lecture Notes in Computer Science, vol 9911. Springer, Cham.

[36] S. Zafeiriou, M. Nicolao, I. Kotsia, F. Benitez-Quiroz, G. Zhao, "Aff-wild: Valence and arousal in-the-wild challenge", *IEEE CVPR Workshop*, 2017.

[37] Sanghyun Woo, Jongchan Park, Joon-Young Lee, and In So Kweon. Cbam: Convolutional block attention module. In Proceedings of the European Conference on Computer Vision (ECCV), pages 3–19, 2018.

[38] D. Meng, X. Peng, K. Wang, Y. Qiao, "Frame attention networks for facial expression recognition in videos.", arXiv preprint arXiv:1907.00193, 2019.

[39] K. Zhu ; Z. Du ; W. Li ; D. Huang ; Y. Wang ; L. Chen, "Discriminative Attention-based Convolutional Neural Network for 3D Facial Expression Recognition," *IEEE FG 2019*, May 2019.

[40] K. Wang, X. Peng, J. Yang, D. Meng, Y. Qiao, (2019). Region Attention Networks for Pose and Occlusion Robust Facial Expression Recognition.

[41] J. M. Haut, M. E. Paoletti, J. Plaza, A. Plaza, J. Li, "Visual attention-driven hyperspectral image classification", *IEEE Trans. Geosci. Remote Sens.*

[42] D. Kollias, S. Zafeiriou: "Expression, Affect, Action Unit Recognition: Aff-Wild2, Multi-Task Learning and ArcFace". BMVC, 2019.

[43] Kollias, Dimitrios, and S. Zafeiriou. "Aff-Wild2: Extending the Aff-Wild Database for Affect Recognition." (2018).

[44] Kollias, Dimitrios, and S. Zafeiriou. "A Multi-Task Learning & Generation Framework: Valence-Arousal, Action Units & Primary Expressions." (2018).

[45] D. Kollias, P. Tzirakis, M. A. Nicolaou, A. Papaioannou, G. Zhao, B. Schuller, I. Kotsia, S. Zafeiriou, "Deep Affect Prediction in-the-wild: Aff-Wild Database and Challenge, Deep Architectures, and Beyond". International Journal of Computer Vision (IJCV), 2019.

[46] D. Kollias, A. Schulc, E. Hajiyev, and S. Zafeiriou, "Analysing affective behavior in the fifirst abaw 2020 competition," 2020.